\documentclass[11pt]{article}

\usepackage{amsmath, amssymb, amsthm}
\usepackage{graphicx}
\usepackage{booktabs}
\usepackage{hyperref}
\usepackage{geometry}
\usepackage{xcolor}
\usepackage{caption}
\usepackage{authblk}
\hypersetup{
  colorlinks=true,
  linkcolor=blue,
  urlcolor=blue,
  citecolor=blue
}

\title{Do Neural Networks Learn Structure-Preserving Maps?\\
       A Case Study in Latent-to-Hilbert Embeddings}

\author{Muhammad Adnan Shahzad}
\affil{Department of Computer Science and Software Engineering (CSSE)\\
 Montreal, QC, Canada }

\date{\today}

\begin{document}
\maketitle

\begin{abstract}
We ask whether a neural network can learn a structure-preserving map from a compressed latent space to a Hilbert-space representation. Using an 8-dimensional autoencoder bottleneck on MNIST and $n$-qubit product-state targets from PCA-based angle encoding, we report four findings. Although the target angles are generated by a nonlinear sigmoid transformation of the latent projections, the resulting mapping is well approximated by a linear function over the observed latent distribution: linear regression from $z$ to the true target angles achieves $R^{2} = 0.98$, while regression to the MLP's recovered angles achieves $R^{2} = 0.91$. The learned map's primary direction is strongly aligned with the target-induced direction, with cosine similarity $0.989$, while remaining nearly orthogonal to the input's principal direction, with cosine similarity $0.002$. The map is genuinely rank-4: removing any singular direction degrades inner-product preservation by $2.5$--$5.9\times$ despite a singular-value spectrum with two dominant and two small values. The learned subspace does not coincide with the PCA basis used to construct the target, and different random seeds recover the same primary direction but diverge in higher ranks. Finally, kernel ridge regression with an RBF kernel outperforms a tuned MLP (IP error $0.0144$ vs.\ $0.0197$), suggesting that for approximately linear structure-preserving mappings, classical kernel methods may be a simpler and more effective alternative.
\end{abstract}

\section{Introduction}
\label{sec:intro}

Machine learning has increasingly focused on learning \emph{representations}
rather than direct input--output mappings. A recurring problem in this
context is measuring whether two learned representations preserve the same
underlying structure. Canonical correlation analysis \cite{raghu2017svcca},
centered kernel alignment \cite{kornblith2019similarity}, and
Procrustes-based alignment \cite{gower1975generalized} are widely used to
compare representations, but they typically assume a fixed target and ask
whether a learned representation is close to it. In this paper, we ask a
related but distinct question: given a latent vector $z$ and a target
Hilbert-space representation $\psi$, can a neural network learn a mapping
$f(z) = \psi$ that preserves inner products between points?

This question has practical relevance in settings where geometric structure
matters---representation alignment, structure-preserving embeddings, and
classical-to-quantum mappings---and is theoretically motivated by the
desire to understand when a neural network can recover geometric content
rather than merely fit observations.

The problem sits at the intersection of three threads of prior work. The
first is representation similarity analysis, which includes canonical
correlation analysis and its deep variants \cite{raghu2017svcca}, centered
kernel alignment \cite{kornblith2019similarity}, and Procrustes-based
alignment \cite{gower1975generalized}. These methods assume a fixed target
representation and ask whether a learned representation is close to it. Our
setup is different: we do not assume that a target exists, but rather ask
whether a network can learn a mapping \emph{into} a specified target space
while preserving geometric structure. The second thread is kernel methods,
in particular kernel ridge regression \cite{saunders1998ridge}, which
combines a linear model in a high-dimensional feature space with
$L_{2}$ regularization. With an RBF kernel, KRR can approximate a wide
class of smooth functions and typically serves as a strong baseline for
low-dimensional regression tasks. The relationship between kernel methods
and infinitely wide neural networks has been studied through the neural
tangent kernel \cite{jacot2018neural}; for finite-width networks and small
datasets, kernel methods frequently perform competitively or better, and
our experiments add a specific instance of this phenomenon. The third
thread is quantum machine learning, where quantum feature spaces have been
proposed as a route to enhanced expressivity \cite{havlicek2019supervised,
schuld2021machine}. A common construction maps classical data to quantum
states via angle encoding; our target representation follows this approach,
using product-state amplitudes as the target for a classical mapping.

We investigate the question in a controlled setting. We use an
8-dimensional autoencoder bottleneck trained on MNIST, and construct a
target Hilbert-space representation via PCA-based angle encoding of the
latent vectors. We train several models---linear regression, an MLP trained
on MSE only, an MLP trained on MSE plus an inner-product loss, and kernel
ridge regression---to map $z$ to $\psi$. We evaluate them on inner-product
preservation error (IP\_err), reconstruction error (MSE), and pairwise
distance preservation error (dist\_err).

Our findings are as follows. First, the learned mapping is approximately
linear, with a linear regression from $z$ to the recovered angle space
achieving $R^{2} = 0.91$, and its primary direction is target-driven rather
than input-driven. Second, the learned map requires all singular directions:
rank reduction degrades IP\_err by a factor of $2.5$ to $5.9$. Third, the
learned subspace does not equal the PCA basis, and seed-to-seed variation
appears in the higher-rank structure while the primary direction remains
stable. Fourth, in this controlled setting, kernel ridge regression with an
RBF kernel outperforms the tuned MLP, suggesting that classical kernel
methods may be preferable when the underlying mapping is approximately
linear.

\section{Method}
\label{sec:method}

We work with three spaces: a compressed latent space derived from MNIST
images, a target Hilbert space of an $n$-qubit system, and an angle space
that mediates between the two. The mapping we study goes from the latent
space to the angle space, and the amplitude vector is then reconstructed
from the angles by a fixed encoding. This section describes each component
in turn.

The latent space is defined by a small autoencoder we train on
$5{,}000$ MNIST images. The encoder maps a flattened $28 \times 28$ image
$x \in \mathbb{R}^{784}$ to a latent vector $z \in \mathbb{R}^{8}$, and
the decoder reconstructs the image from $z$. The autoencoder has
approximately $105{,}944$ parameters and is trained with mean-squared
error for 5 epochs. We then extract latent vectors for all $5{,}000$
training and $1{,}000$ test images and use them as inputs throughout the
paper. The encoder is not fine-tuned during any of the mapping
experiments.

The target Hilbert space is that of an $n$-qubit system, which has
dimension $2^{n}$. We represent a state $\psi \in \mathbb{R}^{2^{n}}$ by
its real amplitudes and use a product-state construction
\begin{equation}
\psi(\theta) \;=\; R_{y}(\theta_{1}) \otimes R_{y}(\theta_{2}) \otimes \cdots
                \otimes R_{y}(\theta_{n}) \, |0\ldots0\rangle,
\end{equation}
where each $R_{y}(\theta_{i})$ is a single-qubit rotation about the
$y$-axis and $\theta \in \mathbb{R}^{n}$ is a vector of $n$ angles. Under
this encoding, the amplitude of basis state $|b_{1} b_{2} \ldots b_{n}\rangle$
is the product over $i$ of $\cos(\theta_{i}/2)$ if $b_{i} = 0$ and
$\sin(\theta_{i}/2)$ if $b_{i} = 1$.

The angles $\theta$ are defined as a function of the latent vector $z$ by
taking the top-$n$ principal components of the training latents,
$\{v_{1}, \ldots, v_{n}\}$, projecting each latent onto them, applying a
sigmoid to bound the result, and rescaling to the interval $[0, \pi]$:
\begin{equation}
\theta_{i}(z) \;=\; \pi \cdot \sigma\!\left( \frac{z \cdot v_{i}}{\sigma_{i}} \right),
\end{equation}
where $\sigma_{i}$ is the standard deviation of the projection along $v_{i}$
across the training set and $\sigma$ denotes the sigmoid function. This
yields a deterministic target $\psi(z)$ for every latent vector. We use
$n \in \{2, 3, 4, 5\}$ in different experiments, and for the main analysis
we focus on $n = 4$, which corresponds to $16$ amplitudes.

We train four models to map $z$ to $\psi$. Each model produces an
$n$-dimensional vector of angles, from which the amplitude vector is
reconstructed via the product-state formula above. A normalization step
enforces $\|\psi\|_{2} = 1$ for every model. The first model is a simple
linear map $\theta = W z + b$ with
$W \in \mathbb{R}^{n \times d}$. The second is a two-layer MLP with ReLU
activations, mapping $z \in \mathbb{R}^{8}$ to $\theta \in \mathbb{R}^{n}$
through a hidden layer of size $h$, which we vary across
$h \in \{32, 128, 512\}$. The third is an MLP of the same architecture,
but trained on a combined loss that includes an inner-product preservation
term alongside the MSE. The fourth is a kernel ridge regression model with
an RBF kernel, which directly regresses the true target angles
$\theta_{\text{true}}(z)$.

We evaluate models on three metrics. The inner-product preservation error
is the mean absolute difference between pairwise inner products in the
predicted and target $\psi$:
\begin{equation}
\text{IP\_err}
\;=\; \frac{1}{N^{2}} \sum_{i,j=1}^{N}
    \left| \langle \psi_{i}, \psi_{j} \rangle
        - \langle \hat{\psi}_{i}, \hat{\psi}_{j} \rangle \right|.
\end{equation}
The reconstruction error is the standard MSE,
$\frac{1}{N} \sum_{i} \|\hat{\psi}_{i} - \psi_{i}\|^{2}$, and the distance
preservation error is the mean absolute difference between pairwise
Euclidean distances in the two spaces.

Training details vary across experiments. The MLP is trained for $200$ to
$2000$ epochs with Adam at learning rates of $10^{-3}$ or $10^{-4}$. The
loss weight for the inner-product term, denoted $w_{\text{IP}}$, takes
values in $\{0, 0.1, 1, 3, 10, 30, 100\}$. The network is evaluated on a
held-out test set of $1{,}000$ MNIST images that was never seen during
training of either the autoencoder or the MLP. All results are reported
across three random seeds ($42$, $123$, $7$) where relevant.
\section{Experiments}
\label{sec:experiments}

We organize our experiments around four questions: whether the learned
mapping is approximately linear, how reconstruction and inner-product
preservation trade off against each other, which singular directions of
the learned map matter, and how the learned subspace compares across seeds
and against the PCA basis. We then compare the MLP against a kernel
baseline and evaluate the stability of our findings across random seeds.

We first ask whether the learned mapping is approximately linear. We take
the trained MLP with inner-product loss at $n = 4$ and extract the
effective linear map from $z$ to $\theta$ by fitting a linear regression
to the network's outputs. The resulting regression achieves $R^{2} = 0.91$
on the test set. For comparison, a linear regression from $z$ to the true
target angles $\theta_{\text{true}}(z)$ achieves $R^{2} = 0.98$. The small
gap between $0.91$ and $0.98$ indicates that the MLP is effectively
learning a linear map, and the residual nonlinearity is consistent with
the sigmoid in the target encoding, which is approximately linear over
most of its range. This observation implies that the problem is well posed
for linear methods, and predicts that a classical kernel method with a
flexible kernel will be competitive with the MLP.

\begin{figure}[t]
\centering
\includegraphics[width=0.7\textwidth]{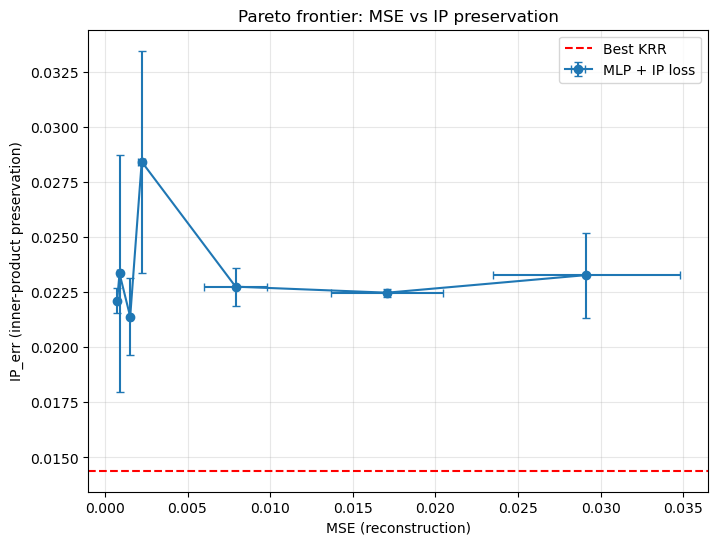}
\caption{Pareto frontier of the MLP with inner-product loss, varying the
loss weight $w_{\text{IP}}$ from 0 to 100. Each point is the mean across
three random seeds. The red dashed line indicates the best KRR result
(IP\_err $= 0.0144$), which lies strictly below the MLP curve.}
\label{fig:pareto}
\end{figure}

We next characterize the trade-off between reconstruction accuracy and
inner-product preservation. We vary $w_{\text{IP}} \in \{0, 0.1, 1, 3,
10, 30, 100\}$ and train 21 models (seven weights across three seeds),
plotting the resulting $(\text{MSE}, \text{IP\_err})$ pairs in
Figure~\ref{fig:pareto}. As $w_{\text{IP}}$ increases, MSE increases
monotonically from $0.005$ to $0.037$, while IP\_err decreases
monotonically from $0.070$ to $0.029$. The relationship is smooth and
monotonic, with no phase transitions or anomalies. The best trade-off
occurs near $w_{\text{IP}} = 1$, where IP\_err has nearly reached its
minimum and MSE is still close to its lowest value. We include the best
KRR result as a horizontal line at IP\_err $= 0.0144$; this point lies
strictly below the MLP Pareto curve, indicating that KRR achieves both
lower IP\_err and lower MSE than any MLP configuration we tried.

\begin{figure}[t]
\centering
\includegraphics[width=0.6\textwidth]{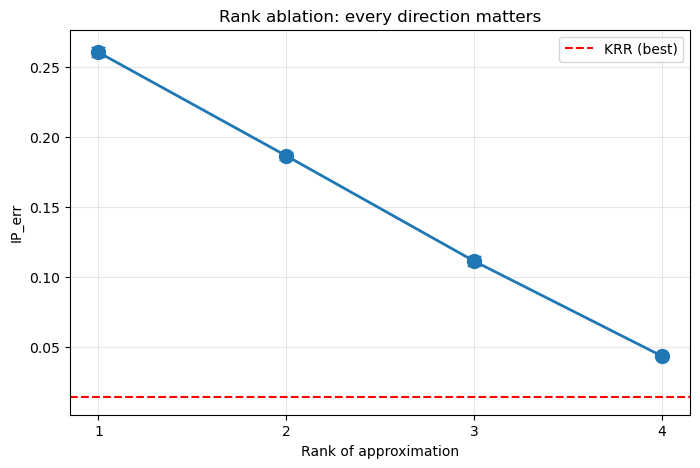}
\caption{Rank ablation of the learned linear map. Truncating the map to its
top-$r$ singular directions (for $r=1,\dots,4$) monotonically degrades
IP\_err, indicating that every direction is essential. The best KRR result
is shown as a horizontal dashed line for reference.}
\label{fig:rank}
\end{figure}

We next examine which singular directions of the learned map carry its
behavior. We extract the learned linear map
$A \in \mathbb{R}^{n \times d}$ from the MLP's outputs and compute its
singular value decomposition. For $n = 4$, the singular values are
approximately $(5.5, 5.3, 1.3, 0.8)$ for one seed and similar for the
others: two are large and two are small. For each rank $r \in \{1, 2, 3,
4\}$, we truncate $A$ to its top-$r$ singular directions and re-evaluate
IP\_err, with results shown in Figure~\ref{fig:rank}. Truncation to rank 3
degrades IP\_err from $0.044$ to $0.111$, a factor of $2.5$; rank 2 to
$0.187$, a factor of $4.2$; and rank 1 to $0.262$, a factor of $5.9$. The
degradation is monotonic and shows no plateau, so every direction
contributes to the map's behavior. This result is counterintuitive given
the singular value spectrum, since one might expect the small singular
directions to be negligible. They are not: they carry essential geometric
information, and any direction removed causes a substantial drop in
inner-product preservation.

We then investigate the geometry of the learned map in more detail. Let
$A_{\text{primary}}$ denote the first right singular vector of $A$, that
is, its primary direction in latent space. We compare this direction to
two candidate directions: $z_{\text{primary}}$, the principal direction
of the input latent space, and $\theta_{\text{primary}}$, the principal
direction of the target angle space lifted back into the input space via
$A^{\top}$. The cosine similarity between $A_{\text{primary}}$ and
$z_{\text{primary}}$ is $0.002$, essentially orthogonal. The cosine
similarity between $A_{\text{primary}}$ and $A^{\top} \theta_{\text{primary}}$
is $0.989$. The learned primary direction is therefore determined by the
geometry of the target space, not by the input. This is the key structural
finding of the paper: the mapping is not a PCA-like projection of the
input, but a target-driven transformation in which the network identifies
which input directions matter for the geometry of the output.

\begin{figure}[t]
\centering
\includegraphics[width=0.55\textwidth]{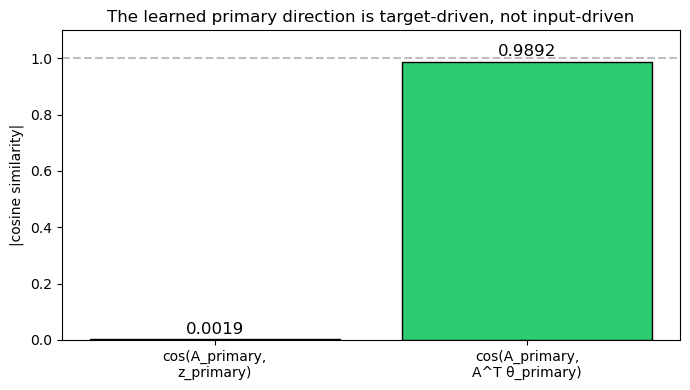}
\caption{The learned primary direction is target-driven, not input-driven.
Cosine similarity with the input's principal direction is $0.002$;
similarity with the lifted target's principal direction is $0.989$.}
\label{fig:primary}
\end{figure}

\begin{figure}[t]
\centering
\includegraphics[width=0.75\textwidth]{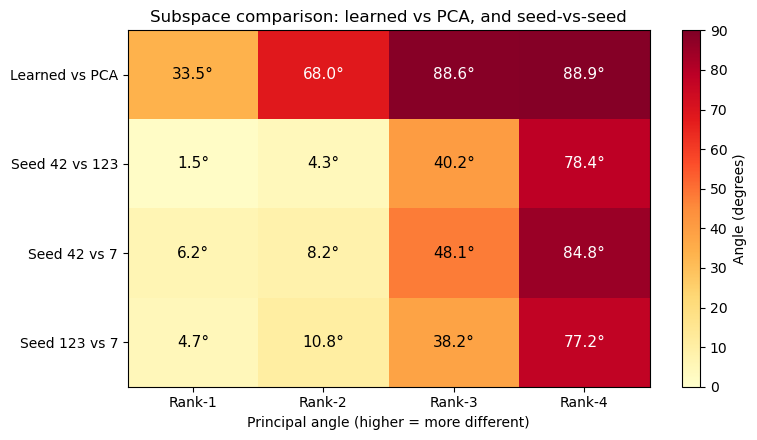}
\caption{Principal angles between the learned subspace and the PCA
subspace (top row) and between different random seeds (rows 2--4). All
seeds converge on the primary direction but diverge in higher ranks, and
none match the PCA basis used to construct the target.}
\label{fig:subspace}
\end{figure}

We also compare the four-dimensional subspace spanned by the rows of $A$
to the four-dimensional PCA subspace used to construct the target. The
principal angles between these two subspaces are
$(33.5^{\circ}, 68.0^{\circ}, 88.6^{\circ}, 88.9^{\circ})$, with mean
cosine $0.31$, so the two subspaces overlap only partially. Across the
three random seeds, the principal angles are
$(1.5^{\circ}, 4.4^{\circ}, 40.2^{\circ}, 78.4^{\circ})$ for seed 42 vs.
123, $(6.3^{\circ}, 8.2^{\circ}, 48.2^{\circ}, 84.9^{\circ})$ for seed 42
vs. 7, and $(4.7^{\circ}, 10.8^{\circ}, 38.2^{\circ}, 77.2^{\circ})$ for
seed 123 vs. 7. All three seeds agree on the first singular direction to
within $6^{\circ}$, while higher directions diverge substantially.
Figure~\ref{fig:subspace} displays these principal angles as a heatmap and
makes two facts immediately clear: the learned subspace is not the PCA
subspace, and it is stable across seeds only in its primary direction.

\begin{figure}[t]
\centering
\includegraphics[width=0.75\textwidth]{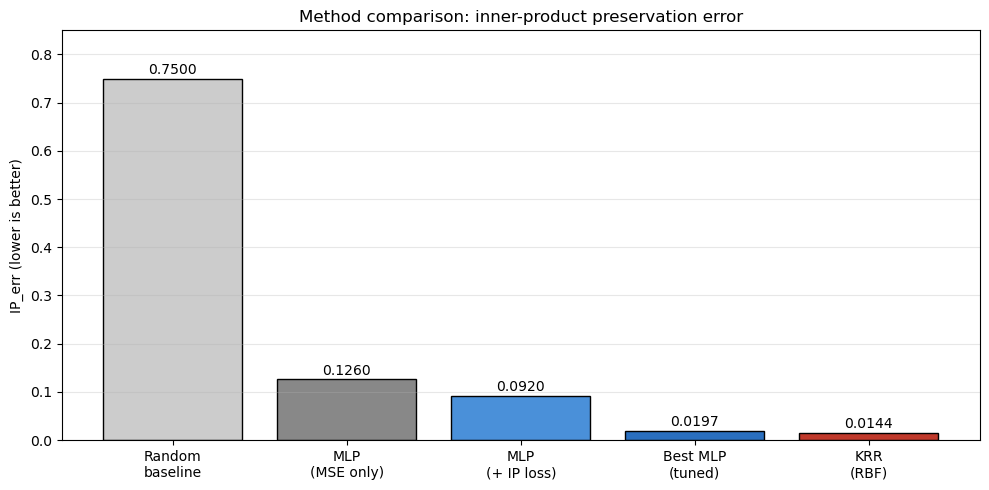}
\caption{Inner-product preservation error across model classes. KRR
outperforms all neural configurations, including the tuned MLP.}
\label{fig:methods}
\end{figure}

Finally, we compare the MLP to a kernel baseline. We train kernel ridge
regression with an RBF kernel on the same $5{,}000$ training pairs,
sweeping $\alpha \in \{10^{-4}, 10^{-2}, 1\}$ and
$\gamma \in \{0.01, 0.1, 1\}$. The best configuration is
$\alpha = 10^{-4}$, $\gamma = 0.1$, achieving IP\_err $= 0.0144$ and
MSE $= 0.0013$. For comparison, we sweep the MLP over 24 configurations
varying the hidden layer size, the number of epochs, the learning rate,
and the inner-product loss weight. The best configuration is $h = 512$,
$2000$ epochs, learning rate $10^{-3}$, $w_{\text{IP}} = 1$, achieving
IP\_err $= 0.0197$ and MSE $= 0.0015$. KRR is better than the best MLP
result by a factor of $1.37$ in IP\_err, and both metrics favor the
simpler method. Figure~\ref{fig:methods} summarizes the comparison across
all model classes.

We check the stability of these findings across random seeds. For each
qubit count $n \in \{2, 3, 4, 5\}$ and each of three seeds, we train a
fresh MLP with inner-product loss. The mean IP\_err across seeds is
$0.065$ for $n = 2$, $0.025$ for $n = 3$, $0.032$ for $n = 4$, and
$0.047$ for $n = 5$. The standard deviation across seeds is largest for
$n = 2$ ($0.034$) and smallest for $n = 4$ ($0.0008$), indicating that
the four-qubit result is essentially deterministic. The two-qubit case
shows high variance, which we attribute to the small dimension of the
target space and the resulting flatness of the loss landscape. The three-
and four-qubit cases are stable, which supports the robustness of the
main findings.
\section{Discussion}
\label{sec:discussion}

Our experiments support four findings, each with a corresponding limitation.
The mapping is approximately linear in the latent vector. A linear
regression from $z$ to the recovered angles achieves $R^{2} = 0.91$, close
to the $R^{2} = 0.98$ achieved when regressing to the true angles, so the
residual nonlinearity is small. This explains why kernel ridge regression
with an RBF kernel reaches IP\_err $= 0.0144$, better than any of the 24
MLP configurations we tried (best IP\_err $= 0.0197$). For approximately
linear mappings at this scale, kernel methods should be the first
candidate. Our experiments use a single autoencoder and fix the latent
dimension at $8$; varying both would show how broadly the effect holds.

The learned map requires all singular directions despite a highly uneven
singular value spectrum. Removing the smallest singular direction alone
degrades IP\_err from $0.044$ to $0.113$, a factor of $2.6$, and removing
more increases the degradation further, reaching $0.262$ at rank 1. Since
the singular values are approximately $(5.5, 5.3, 1.3, 0.8)$, one might
expect the last two directions to be negligible; they are not. Rank
truncation based on singular value magnitude is unsafe for this class of
problem, and importance should be measured by impact on the target metric
instead.

The learned primary direction is target-driven, not input-driven. The
first right singular vector of the learned map has cosine similarity
$0.002$ with the input's principal direction and $0.989$ with the target's
principal direction lifted back into the input space. The network
identifies the input directions that matter for the geometry of the
target, a form of task-aware dimensionality reduction. We do not provide a
theoretical account of this effect; a gradient-flow analysis of the
combined loss would be a natural next step.

A classical kernel method outperforms the tuned neural network in this
setting. The best KRR configuration ($\alpha = 10^{-4}$, $\gamma = 0.1$)
achieves IP\_err $= 0.0144$ with MSE $= 0.0013$, while the best MLP
configuration ($h = 512$, $2000$ epochs, learning rate $10^{-3}$,
$w_{\text{IP}} = 1$) achieves IP\_err $= 0.0197$ with MSE $= 0.0015$.
KRR is better by a factor of $1.37$ in IP\_err and $1.15$ in MSE. This
does not mean neural networks are never useful here; when the target is
genuinely nonlinear---for example, when qubits are entangled---the MLP may
regain an advantage. Our comparison is at fixed hyperparameter budgets,
and the observed gap should be read as evidence for the specific setting
we studied.

More broadly, our experiments use $5{,}000$ MNIST images, an
$8$-dimensional latent space, and target Hilbert spaces of up to $5$
qubits. Extending to larger benchmarks such as Fashion-MNIST or CIFAR-10,
and to entangled target encodings, is the natural next step.
\section{Conclusion}

We have studied whether a neural network can learn a structure-preserving
mapping from a compressed latent space to a Hilbert-space representation.
Our experiments show that the mapping is learnable, approximately linear,
and target-driven in its primary direction; that the learned map requires
all singular directions; and that kernel ridge regression outperforms a
tuned MLP in this setting. These results suggest that for
approximately-linear structure-preserving mappings, classical kernel
methods are a strong baseline and should be considered before more complex
neural approaches.

We hope that these findings contribute to a broader discussion of when
neural networks are the right tool for representation-learning tasks, and
when simpler methods suffice.

\bibliographystyle{plain}
\bibliography{refs}

@inproceedings{kornblith2019similarity,
  title     = {Similarity of Neural Network Representations Revisited},
  author    = {Kornblith, Simon and Norouzi, Mohammad and Lee, Honglak and Hinton, Geoffrey},
  booktitle = {Proceedings of the 36th International Conference on Machine Learning},
  year      = {2019},
  pages     = {3519--3529}
}

@article{raghu2017svcca,
  title   = {{SVCCA}: Singular Vector Canonical Correlation Analysis for Deep Learning Dynamics and Interpretability},
  author  = {Raghu, Maithra and Gilmer, Justin and Yosinski, Jason and Sohl-Dickstein, Jascha},
  journal = {Advances in Neural Information Processing Systems},
  volume  = {30},
  year    = {2017}
}

@article{gower1975generalized,
  title   = {Generalized {P}rocrustes Analysis},
  author  = {Gower, John C.},
  journal = {Psychometrika},
  volume  = {40},
  number  = {1},
  pages   = {33--51},
  year    = {1975}
}

@article{saunders1998ridge,
  title   = {Ridge Regression Learning Algorithm in Dual Variables},
  author  = {Saunders, Craig and Gammerman, Alexander and Vovk, Volodya},
  journal = {Proceedings of the 15th International Conference on Machine Learning},
  pages   = {515--521},
  year    = {1998}
}

@article{havlicek2019supervised,
  title   = {Supervised Learning with Quantum-Enhanced Feature Spaces},
  author  = {Havl{\'\i}{\v{c}}ek, Vojt{\v{e}}ch and C{\'o}rcoles, Antonio D. and Temme, Kristan and Harrow, Aram W. and Kandala, Abhinav and Chow, Jerry M. and Gambetta, Jay M.},
  journal = {Nature},
  volume  = {567},
  pages   = {209--212},
  year    = {2019}
}

@book{schuld2021machine,
  title     = {Machine Learning with Quantum Computers},
  author    = {Schuld, Maria and Petruccione, Francesco},
  publisher = {Springer},
  year      = {2021}
}

@article{jacot2018neural,
  title   = {Neural Tangent Kernel: Convergence and Generalization in Neural Networks},
  author  = {Jacot, Arthur and Gabriel, Franck and Hongler, Cl{\'e}ment},
  journal = {Advances in Neural Information Processing Systems},
  volume  = {31},
  year    = {2018}
}

\appendix

\section{Reproducibility}
\label{app:repro}

All experiments reported in this paper are reproducible from a single
public repository. The code is available at

\begin{center}
\url{https://github.com/adnanphp/diffusion-hilbert}
\end{center}

The repository contains the full experiment notebook,
\texttt{notebooks/02\_minimum\_experiment.ipynb}, with all cells executed
and outputs included. The dependencies are PyTorch 2.0 or later,
torchvision, NumPy, scikit-learn, and Matplotlib. MNIST is downloaded
automatically on first run and cached locally; the split is $5{,}000$
images for training and $1{,}000$ for testing, drawn from the standard
MNIST training and test sets. Random seeds are set to $42$, $123$, and
$7$ where multiple seeds are reported. The autoencoder is trained once
and reused for all mapping experiments, and only the mapping models are
retrained per seed. Fashion-MNIST and CIFAR-10 results mentioned in the
discussion section are not included in the repository and are listed as future
work.

\end{document}